# Probing the Effect of Selection Bias on Generalization: A Thought Experiment

John K. Tsotsos[1]

Electrical Engineering and Computer Science, and
Centre for Innovation in Computing at Lassonde
York University, Toronto, Canada
tsotsos@eecs.yorku.ca

Jun Luo

Noah's Ark Lab,
Huawei Technologies,
Markham, Canada
jun.luo1@huawei.com

## Abstract

Learned systems in the domain of visual recognition and cognition impress in part because even though they are trained with datasets many orders of magnitude smaller than the full population of possible images, they exhibit sufficient generalization to be applicable to new and previously unseen data. Since training data sets typically represent such a small sampling of any domain, the possibility of bias in their composition is very real. But what are the limits of generalization given such bias, and up to what point might it be sufficient for a real problem task? Although many have examined issues regarding generalization from several perspectives, this question may require examining the data itself. Here, we focus on the characteristics of the training data that may play a role. Other disciplines have grappled with these problems also, most interestingly epidemiology, where experimental bias is a critical concern. The range and nature of data biases seen clinically are really quite relatable to learned vision systems. One obvious way to deal with bias is to ensure a large enough training set, but this might be infeasible for many domains. Another approach might be to perform a statistical analysis of the actual training set, to determine if all aspects of the domain are fairly captured. This too is difficult, in part because the full set of important variables might not be known, or perhaps not even knowable. Here, we try a different, simpler, approach in the tradition of the *Thought Experiment,* whose most famous instance is perhaps Schrödinger's Cat, to address part of these problems. There are many types of bias as will be seen, but we focus only on one, selection bias. The point of the thought experiment is not to demonstrate problems with all learned systems. Rather, this might be a simple theoretical tool to probe into bias during data collection to highlight deficiencies that might then deserve extra attention either in data collection or system development.

# 1.0 Introduction

We are motivated in this paper by two observations. First, there seems to be little effort in the machine learning field to routinely validate training data in the sense of ensuring that all important data variations are sufficiently captured statistically and without bias. This deficiency has famously manifested itself in a variety of learned systems that make spectacular errors widely publicized in international media. Second, there is the belief that a system's generalization properties can be ameliorated via manipulations of the given training data, and there is a huge variety of these from simple to very sophisticated (Shorten & Khoshgoftaar, 2019). Many claim that these suffice to remedy any problems arising due to data biases, yet those same problems seem to persist because generalizations cannot deal even with tiny adversarial manipulations (e.g., Xu et al. 2020). Our goals for this paper therefore are first, to begin a discussion of these points and their implications, especially for practical systems where amelioration on its own is not a sufficient goal because there are engineering specifications to be met, and second, to suggest a simple idea for determining which gaps in training data sampling might have important impact on system performance.

It is self-evident that how an artificial agent of any kind perceives its world is critical to how it reacts to events in its world. Within perception, we focus on vision as the primary perceptual source for humans and a capability that requires roughly half of the cerebral cortex to accomplish, thus a major contributor to an agent's overall intelligence. Vision is also a major perceptual source for robots or other artificial agents. We were also concerned that the explosive development of training sets for learned systems is insufficiently rooted in sound statistical science, particularly with respect to bias which for the tiny sizes of datasets (tiny relative to the full population) is likely a much larger problem than is being acknowledged. In other words, it might be that the current empirical methodologies seen in research and applications might benefit from improvement of their data sampling components.

---

[1] Most of this manuscript was developed while the first author was also affiliated with Noah's Ark Lab, Huawei Technologies, Markham, Canada.

It is certainly true that in vision, no current training set fully captures all dimensions of the representation of visual information. This may lead to Selection Bias which results when the selection of data samples leads to a result that is different from what would have been achieved had the entire target population been considered. We run a thought experiment on a real domain of visual objects that we can fully characterize and look at specific gaps in training data and their impact on qualitative performance as well as their impact on the engineering specifications required for acceptable performance in the domain. Our thought experiment points to three conclusions: first, generalization behavior is determined by how well the particular dimensions of the domain are represented during training; second, the utility of generalization is dependent on the acceptable system error; and third, specific visual features of objects, such as pose orientations out of the imaging plane or colours, may not be recoverable if not represented sufficiently during training. Any currently observed generalization in modern deep learning networks may be more the result of coincidental alignments than a true property of the network or its training regime, and whose utility needs to be confirmed with respect to a system's performance specification. Whereas empirical confirmation may be the best method for such verification, it is a time and resource expensive activity. Our *Thought Experiment Probe* approach, coupled with the resulting *Bias Breakdown*, is far less costly and can be very informative as a first step towards understanding the impact of biases. Further, it could be an important supplementary information component for training systems with critical performance requirements.

## 1.1 Current Thinking about Generalization in Learned Systems

Generalizability refers to the performance difference of a model when evaluated on previously seen data (training data) versus data it has never seen before (testing data). Models with poor generalizability have overfitted the training data. A deep learning image classifier is a mathematical function that maps images to classes. Many such classifiers have been applied productively in a variety of domains. A major reason for their success has been that they do not need to be trained with the full domain of the function in order to exhibit strong classification performance for previously unseen input. The reasons underlying this have been studied for some time yet are not well understood. Bartlett & Maass (2003) write that the VC-dimension of a neural net with binary output measures its 'expressiveness'. The derivation of bounds for the VC-dimension for both binary and real-valued neural nets has been challenging. Bounds for the VC-dimension of a neural net provide estimates for the number of random examples that are needed to train a network so that it has good generalization properties (i.e., so that the error of the network on new examples from the same distribution is at most $\varepsilon$, with probability $\geq 1 - \delta$). For a single application these bounds tend to be too large, since they provide such a generalization guarantee for any probability distribution on the examples and for any training algorithm that minimizes disagreement on the training examples. This might seem to be what we seek, but this is not the case. As will become apparent, we are interested in the cases where the training and test distributions differ in their specific detail.

There are many who have written about network generalization and some recent examples follow. Among some of the most cited are Zhang et al. (2016), Keskar et al. (2016) Neyshabur et al. (2017), Kawaguchi et al. (2017), Wu & Zhu (2017) and Novak et al. (2018), and most recently Fetaya et al. (2020). Zhang et al. presented a simple experimental framework for defining and understanding a notion of effective capacity of machine learning models and showed that the effective capacity of several successful architectures is large enough to memorize the training data. They also distinguish optimization from generalization, arguing that formal measures to quantify generalization are still missing. Neyshabur et al. (2017) discuss different candidate complexity measures that might explain generalization in neural networks. They reach no specific conclusions, however, showing how much is unresolved. Fetaya et al. (2020) reveal that it is impossible to guarantee detectability of adversarially perturbed inputs even for near-optimal generative classifiers. Experimentally, they were able to train robust models for MNIST, but robustness completely breaks down on CIFAR10. They suggest that likelihood-based conditional generative models are ineffective for robust classification.

Zhao et al. (2019) examine generalization in deep reinforcement learning and focus on three issues. First, they look at generalization from a deterministic training environment to a noisy and uncertain testing environment. Secondly, assuming one correctly models environmental variability in a training simulation, there is the question of whether an agent learns to generalize to future conditions drawn from the same distribution, or overfits to its specific training experiences. Thirdly, there is the effect on training due to the impossibility of predicting and accurately modeling the environmental conditions and variability an agent might encounter in the real world (i.e., a predictive model should be robust to, for example, camera type, initial agent state, etc.). They show that standard algorithms and architectures generalize poorly in the face of both noise and environmental shift. They propose a new set of generalization metrics as a way of assisting with these problems. It should be pointed out that there is an implicit assumption in this paper, common in most similar works, that the domain of interest can be effectively characterized and parameterized in the first place; this is not possible in general (see also Barbu et al. 2019 and Afifi & Brown 2019 for additional evidence). Others, such as Senhaji et al. (2020), consider multi-task learning and multi-domain learning as a way of providing behaviour that goes beyond a single task or domain while also minimizing the number of needed parameters. However, like all others, the implicit assumptions that a given domain can be 'known' with respect to its parameters and that training sets adequately cover that set of parameters are made. Both assumptions are not always justified.

Bengio & Gingras (1996) consider some of the input variables that are given for each particular training case, and the missing variables differ from case to case, and suggest that the simplest way to deal with this missing data problem is to replace the missing values by their unconditional mean. More recently, Van Buuren (2018) details many methods for imputing missing data and their pitfalls as do Lin & Tsai (2020) who advocate for the development of novel hybrid approaches by combining statistical and machine learning based techniques, the consideration of three evaluation metrics together, and missing data simulation for both training and testing datasets. Śmieja et al. (2018) propose that missing data can be handled by replacing a neuron's response in the first hidden layer by its expected value, thus proposing a neural/unit-level embodiment of Bengio & Gringras's proposal of using the unconditional mean. Chai et al. (2020) fill gaps in a sparsely sampled dataset by fitting curves to the data and interpolating.

García-Laencina et al. (2010) review methods for pattern classification when there is missing data. They group previous research into 4 categories: 1) methods that delete incomplete cases and develop classifiers using only the complete data portion; 2) methods that estimate missing data and learn classifiers using the edited set, i.e., complete data portion and incomplete patterns with imputed values; 3) methods that use model-based procedures, where the data distribution is procedurally modeled, e.g., by expectation–maximization (EM) algorithm; and, 4) methods where missing values are incorporated into the classifier. These seem to imply that one knows that there is a missing feature or data in advance, and this seems an unreasonable assumption in general. Polikar et al. (2010) claim to address the missing feature problem but admit that they draw an equivalence between missing data and missing features. They assume that the complete set of variables in a feature vector is known but that data samples might have missing values for some.

## 1.2  The Issue of Bias in Data

Issues regarding bias in data are common in any discipline that relies on statistical information based on observational data (e.g., genetics, Balding (2006); epidemiology Gordis (2014)). In epidemiology, bias has been defined as any systematic error in the design, conduct or analysis of a study that leads to an erroneous estimate of an exposure's effect on disease. Such bias reflects systematic errors in research methodology. The types of bias include Selection, Detection, Observation, Misclassification, and Recall. These authors highlight that bias is a complex and well-studied topic (see the classic Sackett (1979) paper or in the more recent Grimes & Schulz (2002), among many others). These mostly examine statistical issues in medicine; care is needed if relating these to Machine Learning, computer vision or AI. Nevertheless, much is clearly instructive since that community has struggled with statistical bias for decades in the context of life-and-death decision-making. In other words, there is a very real cost if error is due to bias; sometimes, especially in the published literature, much of computer vision or AI research currently may not fully consider such costs.

Sackett's (1979) highly cited paper provides a catalog of biases. Bias may be introduced at any stage of a research enterprise. He gives the following high level types (in his words):
   i.  In reading-up on the field (5)
   ii.  In specifying and selecting the study sample (22)
   iii.  In executing the experimental manoeuvre (or exposure) (5)
   iv. In measuring exposures and outcomes (13)

v. In analyzing the data (5)
vi. In interpreting the analysis (5)
vii. In publishing the results.

For 6 of these 7 categories, he gives sub-classes, their number appearing in parentheses after each in the above list. In sum, he points to a total of 56 types of bias that might impact a research effort. The use of the term 'exposure' in the above is appropriate for medicine (as in exposure to a virus, or to a drug). In computer vison or AI, the 'experimental manoeuver' corresponds to the attempt by the computer system to solve or perform a task.

Throughout the history of AI research, including all of its sub-fields, has this level of analysis with respect to bias ever been considered? Not really. Should it be now? Yes, especially as applications are beginning to touch safety-sensitive domains. In the current presentation, we focused primarily on the second item in Sackett's list, bias in specifying and setting up the study sample. Even within this category he lists 22 sub-classes. It is not germane to our argument to list all here and the interested reader is encouraged to seek out the original paper. It is very useful however, to give a few for which it is more straightforward to cast their characteristics from medicine into the computer vision or AI domain. Here are 5 of the 22, in his words (in italics) followed by a possible interpretation relevant for AI:

*Popularity bias: The admission of patients to some practices, institutions or procedures (surgery, autopsy) is influenced by the interest stirred up by the presenting condition and its possible causes.*
   In other words, if many researchers, companies or governments are interested in a particular type of image, (e.g., challenges sponsored by a conference or by a prize, the financial lure of start-ups and funding competitions, etc.) the more likely are samples of that type (or studies of this type as a whole) are to be prioritized and others excluded.

*Wrong sample size bias: Samples which are too small can prove nothing: samples which are too large can prove anything.*
   How does one determine sample population size for a given task and domain? Most in the field (perhaps all) follow the maxim 'the more data the better' (e.g., Anderson 2008) with little regard to its suitability for the desired outcome.

*Missing clinical data bias: Missing clinical data may be missing because they are normal, negative, never measured. or measured but never recorded.*
   Tsotsos et al. (2019) show how the range of possible camera parameters are not well-sampled in modern datasets and how their setting makes a difference in performance. Similarly, datasets are not typically annotated with imaging geometry or lighting characteristics for each image. Yet it is clear each affects an image. Finally, there are many cases of particular data being missing, and several research efforts to ameliorate this problem were given in Section 1.1 above.

*Diagnostic purity bias: When 'pure' diagnostic groups exclude co-morbidity they may become non-representative.*
   The computer vison counterpart here reflects what Marr (1982) prescribed in his theory, that it was designed for stimuli where target and background have a clear psychophysical boundary. That is, object context, clutter, occlusions, lighting effects that somehow prevent a clear figure-ground segregation are not included in his theory. Many research works to this day still insist on images of this kind even though it is clear they represent only a small subset of all realistic images.

*Unacceptable disease bias: When disorders are socially unacceptable (V.D., suicide. insanity) they tend to be under-reported.*
   In modern computer vision, 'nuisance' or 'adversarial' situations are under-reported or avoided by engineering around them, while others suggest they can be ignored.

These are just 5 of the 22 sub-categories. The point is to highlight that bias in how one decides on training sets is a well-known issue and has been studied for decades. The reader may disagree with our specific interpretations (and we invite better interpretations) but the existence of these potential sources of bias are well-accepted and undeniable. Acknowledging and incorporating this thinking into machine learning can only benefit the community and the applicability of the results to real-world, safety-critical problems.

The remainder of this paper will consider Selection Bias in general and propose a theoretical framework for exploring how the impact of specific training data 'holes' might impact generalization with respect to how it measures up to a required performance specification.

## 1.3 Setting Up the Problem

None of the approaches in Section 1.1 seem to have made a definitive difference on the overall issue. None ask the question *If a network is trained with a biased dataset that misses particular samples corresponding to some defining domain attribute, can it generalize to the full domain from which that training dataset was extracted?* It is important to note that due to the impossibly large space of input it is not feasible to empirically prove, using exhaustive testing, that generalization abilities are complete for any network. Perhaps a different perspective might be helpful.

For our purposes, we need the following. First, we note that our focus is on visible light images taken with conventional camera systems. Although the conclusions might apply to other kinds of images (LiDAR, thermal, etc.), these are not considered. Other domains are also not considered, such as NLP, but a similar analysis might be as applicable.

Second, the kinds of generalization errors the above papers report, let alone the empirical error, seem to not be in the kinds of ranges one might expect for high performance engineering tasks. If a system has a maximum empirical error that suffices to pass an engineering specification, then generalization error should also be at that same specification. We will perform our analysis with this as a key consideration.

Third, we note that any image training or test set is a subset of the set of all possible discernable images. This might be trivially obvious, but there is an important point here. Consider images of size 32 x 32 pixels with 3 bits per pixel (one bit per color). The number of possible images of such size is $2^{3072}$ or about $10^{300}$. The number of images with three bytes per pixel and 256 by 256 size is about $10^{500,000}$. The trouble with such estimates is that most of these images do not represent scenes that could possibly have been captured by a camera. Pavlidis (2009), who published the just-mentioned example, concludes that it is impractical to construct training or testing sets of images that cover a substantial subset of all possible images. He estimates that $5^{36}$ is a lower bound on the number of discernible images ($5^{36}$ is about $1.5 \times 10^{25}$). Given this, even ImageNet whose April 30, 2010 size is documented at 14,197,122 images, is 18 orders of magnitude too small. For a particular domain (e.g., face classification), the size of all possible discernable images that contain faces is much, much smaller, but still must be quite large. Making assumptions about face location (e.g., is it centre in the image?), face pose (e.g., is it a full frontal head shot? does the head tilt?), image background (e.g., is it on a plain background?), etc., reduces the domain significantly but do they do so sufficiently? Biases due to other factors are well-documented to cause problems (skin tone, lighting, hair, glasses, masks, etc.). The conclusion here is that the domain of images, even if application-specific, can still contain far more instances than practical to represent or use. More importantly, the set of *attributes* that fully define the domain may be unknown (or unknowable).

Even though there are many effective approaches to creating a training set, bias seems not well-handled. It seems quite easy to not include one or more of a domain's attributes or to not have sufficient representation for a particular one within a training set. This is particularly true for 'in the wild' efforts. Missing data would mean that for some domain attribute, its value in a particular training sample is missing (the color of the traffic light is missing, there is no cloudy scene, etc.). No learning method can possibly know in advance how many attributes define a given domain, nor can it know if enough training data is provided to sufficiently enable a high-quality feature extraction process. For example, suppose it is possible to collect a large number of images of several classes of objects in natural scenes but with the light source always in the same position (i.e., identical imaging geometry, such as if all images are taken on clear, sunny days at around 9am). Would a network trained on this dataset generalize to images of those same objects in the same scenes but with differing imaging geometries (e.g., different light source position with respect to the scene and camera)? What about different camera parameter settings? There already is at least one set of studies to show that camera settings make a very large difference to algorithm performance, both classical algorithms and deep learning algorithms (Andreopoulos & Tsotsos, 2011; Wu & Tsotsos 2017; Tsotsos et al. 2019). The initial training set could be considered biased with respect to the dimension of scene lighting since it did not provide the network with samples of many different variations. Similar situations can be considered for a large variety of other image variations (object size, object, color, object 3D pose, contrast with background, etc.). It is very difficult to think that training sets can be easily defined that provide sufficient training samples along each possible variation. Any practical training process for a non-trivial and not artificially constrained domain is likely biased.

As noted earlier, the types of bias include Selection, Detection, Observation, Misclassification, and Recall bias. For our purposes, the type of bias we focus on is Selection bias (other forms of bias remain for future research). Selection bias can result when the selection of subjects leads to a result that is different from what would have been achieved had the entire target population been considered. In vision, a training set is a very tiny subset of the full population of all possible images. The problem is worse when one considers the feature level because of the combinatorially large number of feature combinations (e.g., Tsotsos 1989). Ahmad & Tresp (1993) also point out this combinatorial nature and conclude that for each of these combinations, enough patterns must be included to accurately estimate the posterior density. The satisfaction of this may not be possible in practice unless the domain is small and very well-defined. This is where the generalization properties observed in learned networks are hoped to help. However, the limitations of this hope are not well understood.

## 1.3 The Approach

Our main focus will be on the effect of domain attributes that are unrepresented in training sets due to the potential intractability of complete representations as mentioned above (or perhaps for other reasons as well). This includes any training dataset where for some defining attribute of the domain of interest, one or more of its possible values has no instance (e.g., if color is an attribute then 'red' is a possible value but there is no instance of it in the training set). To address this problem empirically seems an intractable problem. We thus approach this in the tradition of the *Thought Experiment,* whose most famous instance is perhaps Schrödinger's Cat. Our motivating question was given in the abstract: What are the limits of generalization given such bias, and up to what point might it be sufficient for a real problem task? To be more specific, *If a network is trained with a dataset that misses particular values of some defining domain attribute, can it generalize to the full domain from which that training dataset was extracted while maintaining its performance accuracy?* In vision, no current training set fully captures all properties of the representation of visual information[2]. We run this thought experiment on a real domain of visual objects, LEGO® bricks[3], that we can fully characterize and look at specific gaps in training data and their impact on qualitative performance as well as their impact on the engineering specifications required for acceptable performance in the domain. We reach the following conclusions: first, that generalization behavior is dependent on how sufficiently the particular dimensions of the domain are represented during training; second, that the utility of any generalization is completely dependent on the acceptable system error; and third, that specific visual features of objects, such as pose orientations out of the imaging plane or colours, may not be recoverable if not represented sufficiently during training.

We proceed as follows. We will define a hypothetical task in a real problem domain, the domain of LEGO® bricks. We will then assume the task can be sufficiently solved by a correctly trained neural network. The training dataset will then be manipulated, removing particular attribute values and the assumption will be made that the network now trained with the altered dataset will generalize so that it performs as well as the original. The resulting network will be examined in order to determine how the generalization might come about. Specifically, the approach is *Reductio ad absurdum* via *Backcasting*, and this will be further explained below. Our analysis will tie the feasibility of generalization for particular domain attributes to the target performance specifications, because for any applied domain, this seems a critical consideration. First, we specify our hypothetical problem domain and task.

## 2.0 The Domain of LEGO® Bricks

Suppose the well-known toy manufacturer, LEGO®, wishes to take advantage of computer vision advances to open a new premium service that allows customers to order any combination of bricks in their inventory. They wish to visually recognize LEGO® bricks as they pass in front of a camera on a conveyor belt in order to sort them into bins depending on a specified definition (e.g., by color or by shape, or by number of studs). The full variety of pieces is considered including many special purpose ones as the figures below will show. The bin definitions may vary depending on customer requests. A customer might order 10,000 LEGO® bricks all of a particular color but randomly selected otherwise. Another might wish all bricks that are of the 'plate' category with fewer than 12 studs, but of all colors. In order to visually recognize the bricks, we will use a deep learning approach; let's assume the learning system we employ works perfectly and that training error equals zero. We also assume that the standard procedure to reach the best generalization is followed, as described in many of the papers cited earlier. It must be stressed that we do not seek a solution to this task (which would likely be straightforward). Rather, we use this scenario as the backdrop of a theoretical probe into generalization.

---

[2] Many datasets have quite broad coverage but we are unaware of any that cover the full range of possible direct and ambient illuminations or viewpoints, for example, for any particular domain.

[3] LEGO® played no role in this research, and nothing in this manuscript can be considered as representing any aspect of the company or its products. The choice of these building blocks, as opposed to others, is due to their familiarity around the world as well as to the extensive public documentation provided by many authors on the WWW.

## 2.1 The LEGO® Bricks World

What is the dimensionality of the LEGO® bricks domain? LEGO® bricks can be fully characterized by the following: color, size, type, sub-type, dimension (m x n) in studs, and part attributes. Fortunately, others have put much effort in documenting this domain. An online source of container labels that gives a very broad coverage of bricks that have actually been produced over the lifetime of LEGO® is given by Boen (2018)[4]. The full catalog of the bricks presented there is:

| Category | Number of Sub-Types in Each Category |
|---|---|
| Plates 1xN | 48 |
| Plates 2xN | 49 |
| Plates 3xN-4xN | 25 |
| Bricks 1xN | 48 |
| Bricks 2xN | 33 |
| Wedge - Plates | 29 |
| Tiles | 49 |
| Slopes | 106 |
| Plates 5xN-6xN plus Plates 5xN+ | 22 |
| Hinges | 66 |
| Wedge - Bricks | 45 |

There are also label pages for the Technics® series of bricks (Technic® Bricks, Technic® Beams, Technic® Axles, Technic® Gears, Technic® Pins & Connectors), but these are not included here, and the count reflects only the classic LEGO® collection. It should be noted as will be seen in figures below, that a large assortment of unusual formations is included in these collections; all blocks shown in the following figures are from one of the above categories in Boen's catalog. N in the name of the category in the left column above, refers to the number of studs per brick (e.g., a 2x6 plate is a flat piece with studs placed 2 across and 6 deep). Some of Boen's labels are for bins that store pairs of similar pieces and these are not included in the above count. The counts given are of the unique different brick types in this label dataset; the total number of sub-types is 520. Many pieces do not neatly fit into simple categories, even though they are so included, and examples will appear below.

There is a separate label page of all possible brick colors in the Boen (2018) catalog, Color Bars, and these total 141. Since any brick may have any color, the number of possible unique LEGO® pieces is 520 x 141 = 73,320. This is the space of all possibilities but from a practical standpoint, not all are in current production. Diaz (2008) provides the following estimates:
- LEGO® went from 12,000 different pieces to 6,800 in the last few years, a number that includes the color variations
- Staple colors are red, yellow, blue, green, black and white. We thus assume equal numbers of each of 6 colors - this means 1133 unique brick types.
- Approximately 19 billion LEGO® elements are produced per year. 2.16 million are molded every hour, 36,000 every minute.
- 18 elements in every million produced fail to meet the company's high standards.

---

[4] It should be clear that there are actually several such web sites where LEGO® parts and colors are enumerated and catalogued and they do not all agree in their details. Here, this catalog is chosen but we could have performed the analysis, changed in inconsequential ways, with any of the others.

For the purposes of our argument, this is sufficient to provide a good characterization of the size of the domain and of its dimensions (even though they are only coarsely specified or there are variations). For example, in the Bricks 1xN type, there are 9 simple types, i.e., 1x1, 1x2, 1x3, 1x4, 1x6, 1x8, 1x10, 1x12, and 1x16, but also 40 more complex shapes with varying counts of studs. The number of studs could be used as a dimension on its own, but it would only apply to a subset of all the types. Our argument will not depend on the exact ranges of each dimension.

## 2.2 Detection and Binning Scenario

The following are elements that define the scenario for our problem of sorting bricks into bins depending on a specified bin definition that satisfies customer requests[5]:
- There is a conveyor belt moving at a speed compatible with the speed of production of the bricks, i.e., using the data described earlier, so that 600 elements can be processed per second. It is acceptable that many conveyor belts might be used operating in parallel.
- Lighting is fixed; camera position and optical parameters are fixed and known; the camera's optical axis is perpendicular to the conveyor belt and centred at the belt's centerline (camera is directly overhead and pointing down to the center of the conveyor belt); shadows are minimized or non-existent; the appearance of the conveyor belt permits easy figure-ground segregation.
- An appropriate system exists for understanding a customer request (e.g., "I need 500 bricks, only yellow and with fewer than 6 studs") and deploying the appropriate systems for its realization.
- The pose of each brick on the conveyor belt may be upright, upside down, or on one of its other stable sides. Each brick has its own set of stable sides on which it might lie. Each brick would have at least 2 stable positions perhaps to a maximum of 6 (sides of a cube). We conservatively assume 3 as the average, but this is likely low.
- When on their stable position, bricks may be at any orientation relative to the plane of the conveyor belt; we assume the recognition system understands orientation quantized to 5°, i.e., 72 orientations (360 orientations seems unnecessarily fine-grained whereas 4 seems too coarse. The assumption is made to enable the 'back-of-the-envelope' kind of counting argument made here and could easily be some other sensible value. It also seems sensible due to the restrictions on camera viewpoint described earlier)
- We assume that bricks do not overlap and thus do not occlude each other for the camera.
- The number of possible images using the complete library of bricks can be given by 3 (poses) x 520 (brick types) x 141 (colors) x 72 (orientations) = 15,837,120 (single instance of each part at each orientation and pose). In our argument, we will use the actual production numbers described earlier, namely, 3 (poses) x 6800 (bricks of the standard 6 colors) x 72 (orientations) = 1,468,800.

Examples of user requests could be 'all slopes', 'all 1x4 plates', 'all bricks of color C' or 'all bricks with more than 4 studs', or other similar requests. There may be more than one task executed concurrently, but this is not relevant to our argument. We further assume that the error for this binning task needs to be very small. LEGO® brick manufacturing error is $1.8 \times 10^{-5}$ (from Diaz 2008) and it is assumed that any additional error associated with satisfying these user requests should not add to this error appreciably.

The number of possible images is easily manageable by contemporary infrastructure and also seems to be within the reach of the learning/memorization ability of current neural networks. In other words, a product manager at LEGO® may just demand that such a dataset be generated and that a big enough neural network be used. However, our goal is to use this network as an experimental vehicle to explore generalization and we will do this by investigating how generalization error is affected by deficiencies (selection biases) in training data.

## 3.0 The Thought Experiment: Setup

Thought experiments generally:
- challenge (or even refute) a prevailing theory, often involving proof via contradiction (*Reductio ad Absurdum*);

---

[5] See the construction by Daniel West, https://www.youtube.com/watch?v=04JkdHEX3Yk. As impressive as this is, it does not provide a solution for the task we pose. Nevertheless, he solved a variety of difficult mechanical problems whose solution will be taken as an existence proof for our scenario and not further discussed.

- confirm a prevailing theory;
- establish a new theory;
- simultaneously refute a prevailing theory and establish a new theory through a process of mutual exclusion.

The tactic we will employ here is *Reductio ad Absurdum* via the methodology of *Backcasting* (Robinson 1990) in order to challenge the prevailing belief that deep learning networks generalize in a useful manner and to probe into generalization behavior. Backcasting involves establishing the description of a very definite and very specific future situation. It then involves an imaginary moving backwards in time, step-by-step, in as many stages as are considered necessary, from the future to the present to reveal the mechanism through which that particular specified future could be attained from the present. See Figure 1 for a graphical depiction. Backcasting is not concerned with predicting the future; the major distinguishing characteristic of backcasting analyses is the concern, not with likely futures, but with how desirable futures can be attained.

Assume as the starting point for our backcasting tactic, that the recognition task for the LEGO® brick binning problem can be learned, and that when trained and validated on the full set of samples as defined by the above description, network *A* is produced with empirical error less than $1.8 \times 10^{-5}$ (the overall manufacturing error set by LEGO®). Of course, this might mean that the full set is memorized; and since it is the full set, no generalization is even needed. But this is not relevant here. It does not matter how the network achieves this accuracy, only that it does. We then probe the speculation that if training of the same architecture is biased, its generalization properties will overcome the bias with the same acceptable performance level. We will examine how this speculation might be verified by examining the logical functions that would be needed in the architecture and its performance under that speculation. If we fail the performance specification, this would be the proof by contradiction (*Reductio ad Absurdum)* that our speculation was flawed.

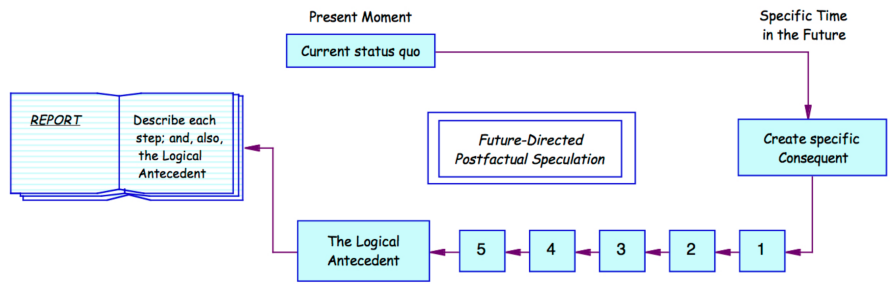

**Figure 1.** Representation of Backcasting (from https://en.wikipedia.org/wiki/Backcasting)

An important assumption is that regardless of the network configuration, the network includes all the classes that would be required to satisfy any user query (classes for brick shape, color, or size defined by number of studs). Supervised learning would include all the required known classes. In other words, the range of possible user queries defines the output classes of the network. We also assume that any user query is parsed correctly. Then, to create the specific consequent required by backcasting, the exact architecture (specifically all output classes required for any user query are kept intact) of this network is re-initialized and re-trained with a biased training set. This hypothetical may seem worrisome, but this is part of the imagination of the experiment. If, for example, a complete re-training of a new network was undertaken with the biased dataset, it might be that one or more output classes corresponding to specifics of user queries would not be present. This would of course be unacceptable for the application, but also would not allow the thought experiment to proceed. Such a scenario reminds of Xu et al. (2019), where they conclude that if labels for a particular category are absent from a dataset, the network will treat the unlabeled object in the image as background or 'other'. In our case, background would mean the conveyor belt and we assume labels also exist.

If a network effectively generalizes, this means that its test error is less than some bound determined by the problem domain. For the LEGO® problem domain, the empirical error bound can be assumed as being less than 0.000018 based on the problem domain manufacturing error described earlier. The goal is to minimize generalization error, the difference between expected error and empirical errors, *G*. As mentioned earlier, the binning process should not exceed such a level of error, and therefore, we will assume $G < 0.000018$ is a reasonable target for overall error.

Assume the newly trained system *A* (the system responsible only for visual recognition of the bricks) satisfies this target; $G_A < 0.000018$. The subscript on *G* will denote the particular network being considered. The main concern is generalization of the visual recognition performance under different training biases. No assumptions are made about any particular capabilities of the network. For example, CNN's are by design somewhat location insensitive. However, all LEGO® blocks are displayed centrally so this is not relevant. No particular knowledge about the task is present in the network before the learning phase.

Why is this a reasonable test? It is a common complaint that datasets for autonomous driving, for example, often are predominantly created in good weather, with no snow or fog or heavy rain or hail, with little traffic, or have other idiosyncrasies that make their datasets represent incomplete training populations. Although the full domain for driving necessarily contains all weather elements, all traffic elements, etc., it is impossible to fully represent these in any finite dataset. Thus, a problem is evident: how can the system architect ensure appropriate generalization behaviour when the training set may not include any samples from a significant portion of the domain population? Is this even feasible? Our thought experiment clearly is as relevant as this. We will explore the issues as our thought experiment unfolds.

# 4.0 The Thought Experiment: Cases Considered

Assume training in our LEGO® domain is done while leaving out one value of some attribute dimension from the training set creating a new network $A^{-i}$ where the *i* signifies the *i*-th dimension where one of its possible values is not represented at all in the training set. This would not be unlike training an autonomous car using a dataset where only sunny, daylight scenes are included. That 'left out' dimension *i*, with respect to visual appearance of the bricks, is one of: color (e.g., leave out one color), size (e.g., leave out all 2x2 bricks), shape (e.g., leave out all slopes), orientation (e.g., leave out images where the long axis of a brick is parallel to the conveyor belt), or pose (e.g., leave out all upside-down poses). We could also consider more than one such 'left out' value or dimension. For each case, we will consider the resulting generalization error.

Going back to Figure 1, the flowchart of the Backcasting process, we use the 'current status quo', namely the state-of-the-art in computer vision, to 'create specific consequent', namely the network $A^{-i}$, which will be assumed to satisfy all production specifications LEGO® would require. We will consider a number of different sources of possible training biases and their impact on error for this network. For each, the overall error will be estimated and given as the sum of the error of the full network *A* ($G_A < 0.000018$) and any new errors introduced as a result of the changes in the training set. As in Figure 1, one or more steps will be taken towards the 'logical antecedent', in other words, towards what must be true in order for network $A^{-i}$ to perform as assumed. Whether or not the logical antecedent is true or even realistic will determine our conclusions regarding generalization.

## 4.1 Training Set Missing One Brick Color

Suppose one color of the staple LEGO® brick set possibilities (red, blue, green, white, black, yellow) is left out of the training dataset and that the resulting system $A^{-c}$ satisfies the empirical error stipulated above. This means 5 colors are sufficiently represented and thus are learned correctly. This places the state of our argument at the 'create specific consequent' stage of the Backcasting flowchart in Figure 1. We now proceed to the boxes in the figure labelled 1 through to the 'logical antecedent'. (These steps will be true for each of the cases described below and not further repeated in those descriptions). Although the color label set referred to above includes 141 colors, these 6 are the standard ones, and in any case, sufficient to demonstrate the point here[6]. If those 5 colors are learned and thus 'in the weights' of the network, then those weights might be invoked when the 'left out' color is seen.

Color is complicated (see Koenderink 2010). The most common representation for images, such as in those datasets in common use, is that each pixel in an image is represented by R (red), G (green) and B (blue) values, each coded using 8 bits. In other words, for each of the the colors, 256 gradations are possible yielding a potential 16,777,216 colors. Luminance or intensity is not independently coded but is rather derived from these as an average of the three values. The values for each of R, B and G come from image sensor outputs, and the most common practice is for the sensor to employ a Bayer filter (a pattern of color filters with specific spatial distribution), and then to be further processed, including to de-mosaic the image. These color filters are designed to specific characteristics; B is a low pass filter, G is a band-pass filter and R is a high-pass filter, attempting to fully cover the range 400-700nm. Spatially, the Bayer filter is arranged to have 2 green filters for each red or blue, that is, each image pixel is the result of these 4 samples, 2 green, one red and one blue. The filter pass ranges overlap only by small amounts. The goal is to match the wavelength absorption characteristics of human retinal cones, which very roughly are: S (blue) 350 - 475nm, peaking at 419nm; M (green) 430 - 625nm, peaking at 531nm, S (red) 460 - 650nm, peaking at 559nm. Rods, for completeness, respond from about 400 - 590nm, peaking at 496nm. An interesting characteristic is that beyond 550nm, only the M and L systems respond. Koenderink points out that the distributions in the human eye are not ideal in how they cover the spectrum; an ideal system would more equally cover the full spectrum. This is what the Bayer filter attempts to accomplish. Of course, there are many color models and encoding schemes, but this is most common.

---

[6] A different author says the 2016 LEGO® Color Palette has 39 solid, 3 metallic, 1 glowing, and 14 translucent colors. Among these, is one red, 2 different yellows, several blues and greens. https://www.bricklink.com/v3/studio/design.page?idModel=169765. We could redo our experiment with red or with both yellows missing and the result would be the same.

Suppose the 'left out' brick color is yellow. As Koenderink points out, the spectra of common objects span most of the visible light spectrum (see his Fig. 5.18 that shows 6 different spectra of real 'yellow' objects; for each, most of the visible spectrum has some representation except blue). So, if one considers the RGB representation across all the bricks, most of the other colored bricks will have some yellow spectral content but this will never be independently represented. The portion of the spectrum we commonly see as yellow is quite narrow, centred around 580nm. In both human vision and using the Bayer filters, these wavelengths are not sampled by the blue component at all. In human vision, the absorption spectra of the L and M cones have good sensitivity to the yellow wavelengths (their peaks are at 531 and 559 respectively). However, using Bayer filters, most applications try to minimize overlap so at the yellow wavelengths, sensitivity is typically low, less than 50% of the peak.

The question we ask is, if no yellow brick is part of the development of $A^{-c}$, what is the resulting generalization error? Since color is 'in the weights' of this network, and all non-yellow objects likely exhibit some yellow in their reflectance spectra, we now ask if it is enough to enable classification of a yellow brick via mixing.

The color humans perceive is a complex function of the spectrum of the illuminant (and ambient reflected spectra), the angle of incidence of the illuminant on the surface being observed, the albedo of the surface, the angle of the viewer with respect to the surface, and the surrounding colors of the surface point being observed, Most of this is captured by the well-known Bidirectional Reflectance Distribution Function (BDRF) (see Koenderink 2010 for more detail). In our brick binning task, much of this can be ignored. The illuminant, its angle to the conveyor belt, the camera and its angle to the conveyor belt, the surrounding surface are all constant. The variables remaining are the surface albedo and the angle between the camera's optical axis and the brick surfaces being imaged.

It has long been accepted that colors can be formed by additive mixtures of other primary colours[7]. For our purposes, this means that even though some spectrum of color is 'in the weights' of the network, their combination cannot necessarily result in any other color if a primary color is missing. Color theory for mixtures tells us:
- green is created by combining equal parts of blue and yellow;
- black can be made with equal parts red, yellow, and blue, or blue and orange, red and green, or yellow and purple[8];
- white can be created by mixing red, green and blue; alternatively, a yellow (say, 580nm) and a blue (420nm) will also give white.

However, red, blue and yellow are primary colors and cannot be composed as a mixture of other colors. If a primary color is unseen during training, there can be no set of weights that would represent it as a combination of other colors. This would mean that if a primary color is unseen, it could not be classified. In a brick classification task such as ours, color is an important dimension because it divides the entire population into 6 groups; then within each group are the different block types.

---

[7] Koenderink provides a history with some emphasis on Grassman's laws from 1853. Bouma in 1947 writes an interpretation of these as: "If we select three suitable spectral distributions (three kinds of light) we can reproduce each color completely by additive mixing of these three basic colors (also called primary colors). The desired result can only be attained by one particular proportion of the quantities of the primary colors".

[8] The author of https://brickarchitect.com/color/ points out that the LEGO Black is not a true black, but rather a very dark gray, and LEGO® White is actually a light orange-ish gray.

The above are all additive mixing rules. Subtractive mixing should be considered as well because networks employ both positive and negative weights, and it might be that this is an alternate avenue for dealing with the color. The additive rules arise by using the RGB color model while the subtractive rules are within the CMYK color model, not often seen in neural network formulations. The three primary colors typically used in subtractive color mixing systems are cyan, magenta and yellow. Cyan is composed of equal amounts of green and light blue. Magenta is composed of equal amounts of red and blue. Yellow is primary and cannot be composed of other colors in both color models. In subtractive mixing, the absence of color is white and the presence of all three primary colors makes a neutral dark gray or black. Each of colors may be constructed as follows:
- red is created by mixing only magenta and yellow
- green is created by mixing only cyan and yellow
- blue is created by mixing only cyan and magenta
- black can be approximated by mixing cyan, magenta, and yellow, but pure black is nearly impossible to achieve.

As Koenderink (2010) says, when describing his Fig. 5.18, the best yellow paints scatter all the wavelengths except the shorter ones. In fact, the spectra he shows for 'lemon skin', 'buttercup flower', 'yellow marigold' or 'yellow delicious apple' have significant strength throughout the red and green regions. In practice, a strong yellow, such as in a brick, could appear as the triple (r, g, b) in an RGB representation, where $0 \leq r, b, g \leq 1$, and r, g $\gg$ b, b being close to 0. It would be reasonable to think that since no yellow brick is seen in training, that there would be no corresponding ability to classify it.

With some generous assumptions about how the colors are represented in the weights and how they combine through the network, there may be a route to combinations that lead to most colors. After all, this would reflect the distributed representations underlying such networks (Hinton 1984): each entity is represented by a pattern of activity distributed over many computing elements, and each computing element is involved in representing many different entities. But color theory tells us that no combinations can yield yellow; it cannot be overcome using a learning strategy that never sees samples of these colors nor by distributed representation strategies. The only conclusion possible is that network $A^{-c}$ where c ∈ (red, blue, yellow) will not properly generalize, but that c ∈ (green, white, black) might. This would of course be unacceptable to LEGO®. On the other hand, a dataset that is biased towards this latter color subset might actually exhibit performance metrics for test error that could appear promising; this would, however, be a false indication. An analysis of failure instances would reveal this problem.

There are many other color spaces. Rasouli & Tsotsos (2017) review 20 different spaces and show how each leads to different characteristics for the detectability of objects. Above, we show only two of these. It might be that the correct choice of color space for particular training data sets leads to different generalization properties for learned systems. This requires further research but the methods described here might be helpful.

Our hypothetical network $A^{-c}$ where $c$ ∈ (red, blue, yellow) would thus exhibit the following error. Leaving out one of these colors, if we assume all LEGO® bricks are made in all colors equally, means that 1/6th of all bricks (since 1/6th of all bricks cannot have their color correctly classified) are erroneously binned. Since the total number of bricks possible is 1,468,800. as enumerated in Section 2.2, a 1/6th error (244800/1468800 = 0.166667) overwhelms the manufacturing error of 0.000018. The overall error, as defined in Section 4.0, of G can be estimated as $G_{A^{-c}} < 0.166685$. For $A^{-c}$ where c ∈ (green, black, white), using generous training assumptions, we can assume no additional error so $G_{A^{-c}} < 0.000018$.

## 4.2 Training Set Missing One Brick Size

Let us name the learned system where one of the sizes is left out of training $A^{-z}$, $z$ ∈ ($s_1, s_2, ....s_k$), the set of all block sizes, and that it satisfies the empirical error stipulated above. There are many sizes in the label set we are using and not all sizes apply to all brick types. The majority of the specialty bricks are of a unique size. Common pieces have several sizes. For example, in the Bricks 1xN type, there are 9 simple sizes, i.e., 1x1, 1x2, 1x3, 1x4, 1x6, 1x8, 1x10, 1x12, and 1x16, but also 40 more complex shapes with varying counts of studs, thus of differing physical size when compared to other bricks but without variation of size with respect to that particular type of brick. In other words, there seem to be at least two different dimensions along which size might be represented: stud count and brick volume. There may be additional ways as well; let's consider stud count only (in Figure 2, the bricks in panels a, b, c, and d have 4, 1, 2, and 3 studs respectively). The minimum stud count is 1 and the maximum is 54 in the label set referenced above. The accurate size distribution is tedious to enumerate; it will likely be as informative to assume that there are equal numbers of each stud count, that is, approximately 73,320/54 = 1358 using the full brick set, or 1133/54 = 21 using the production values for unique brick types cited above. Suppose one stud size is left out of the training set but that the resulting system $A^{-z}$ satisfies the empirical error stipulated above. How could the system generalize to that missing stud count? One might imagine that if a simple linear piece with 4 studs is left out of the training data, with generous assumptions, that the learned portions of the network for the similar shapes might jointly fire and fill in the gap. This would mean that there is some combination of network elements that form a 4-stud straight piece, as shown in Figure 2a. Straightforward possibilities include Figure 2b, c, and d. But then could the 3-stud piece in Figure 2e participate in this composition? How could the 4-stud bricks of Figure 2e-h be made? It is assumed that all the pieces in this figure, and in all other figures, are part of the training set for the ideal network $A$ since they are all art of the brick sets enumerated in Section 2.1.

There are many similar questions given the variety of ways LEGO® has found to make bricks with 4 studs. It seems that the composition with learned pieces might provide a partial answer, but not a complete solution. Certainly, any error measure would be increased even if we assume a partial solution.

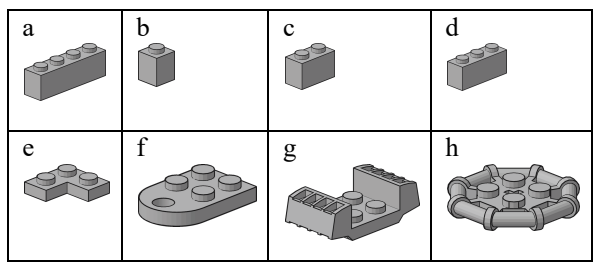

**Figure 2.** Example bricks for Section 4.2.

Our hypothetical network $A^{-z}$ would exhibit the following worst-case error. If all the bricks of a single stud measure are binned incorrectly, that is, 21 out of 1130 bricks, the error is 21/1130= 0.018584 which when summed to the overall production error gives a cumulative error of $G_{A^{-z}} < 0.018602$.

## 4.3 Training Set Missing One Brick Orientation

Suppose one orientation in the imaging plane (i.e., parallel with the conveyor belt - recall the imaging geometry described in Section 2.2) is left out of the training set and that the resulting system $A^{-o}$ satisfies the empirical error stipulated above. Simple data augmentation methods, such as spatial shifts or within the image plane rotations, are likely to permit generalization to one, or perhaps, more orientations being omitted from a training set (more on data augmentation methods below). The lack of good representation of all orientations is probably not an insurmountable problem; this would, however, also depend on whether or not there is a need for precision grasping if a robotic manipulator is used and this is not further considered here. Data augmentation is a relevant method for reducing orientation-in-the-imaging-plane sampling biases in training data given the restricted imaging geometry. The hypothetical network $A^{-o}$ exhibits an overall error that is not increased due to this bias, and thus we can assume that $G_{A^{-o}} < 0.000018$.

## 4.4 Training Set Missing One Brick 3D Pose

Suppose one 3D pose is left out and that the resulting system $A^{-p}$ satisfies the empirical error stipulated above. The LEGO® brick domain is well-suited to an aspect graph representation and this is yet another advantage of choosing these bricks as our domain of interest. We could not easily enumerate all the characteristics of most other domains, certainly not of natural images. First, a brief overview of aspect graphs (Cutler 2003).

An *Aspect* is the topological appearance of an object when seen from a specific viewpoint. Imagine a sphere where each point on the sphere represents the viewing direction formed by that point and the sphere's centre, as is shown in

Figure 3. A *Viewpoint Space Partition* (VSP) is a partition of viewpoint space into maximal regions of constant aspect. An *Event* is a change in topological appearance as viewpoint changes. A constant aspect then is a contiguous region on the sphere that is bounded by changes in topological appearance, i.e., events. The full space partition would show a different region for each face. Consider the simple example of a cube. The separate regions of the space partition would be composed of regions where only one side is visible, only 2 sides are visible and only 3 sides are visible. There is no partition where 4 sides can be visible. There are 26 of these regions. An *Aspect Graph* is a graph with a node for every aspect and edges connecting adjacent aspects. The dual of a viewpoint space partition is the aspect graph. Aspect graphs can be made for convex polyhedra, non-convex polyhedra, general polyhedral scenes (same as the non-convex polyhedra case), and non-polyhedral objects (e.g., torus). In general, for convex objects, the size of the VSP and the aspect graph are of order $O(n^3)$ while for non-convex objects, the VSP and aspect graph are of order $O(n^9)$ under perspective projection ($O(n^2)$ and $O(n^6)$ under orthographic projection respectively, where *n* is the number of aspects) (Plantinga and Dyer 1990). The cube has fewer than this general number because it has 3 pairs of parallel planes that do not intersect and thus do not form a viewing possibility. Recognition algorithms that employ aspect graphs typically match a set of aspects to a possible reconstruction of a hypothesized object (see Rosenfeld 1987 for the first of these; Dickinson et al. 1992 for a nice development using object primitives).

We will assume orthographic projection because we can engineer the imaging geometry to satisfy its properties. If one 3D pose is left out of the training set representation, for example, there are no images of any LEGO® brick with its top surface facing the conveyor belt, then the following results. If one side is never seen, it affects all aspects in which it participates. For a cube, this would mean: the aspect of itself, the aspects with one neighbouring face (4), and the aspects with 2 neighbouring faces (4), for a total of 9 aspects. In other words, 9 of the 26 aspects of the brick need to be generalized using the remaining 17 learned aspects. In other words, 35% (9/26) of the possible constellation of aspects required for recognition would not be available. Recognition would fail if the observed viewpoint led to a set of visible aspects that overlapped these 9 aspects. For bricks more complex than a cube, this would differ.

However, each possible missing pose does not necessarily represent a stable configuration for a brick lying on a conveyor belt. Of the brick types listed above, wedge-plates, tiles, bricks 1xN, slopes, plates 3xN/4xN, plates 2xN, plates 1xN, many would have only 2 stable configurations, right-side up and upside down with no possibility of a stable side-ways configuration.

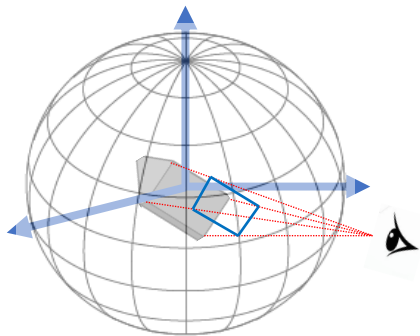

**Figure 3.** Example of a typical LEGO® element - an elongated triangle block - within an aspect representation. The eye is in an approximate position from where the view of the block would be only a rectangle. The aspect seen is shown in blue outline, the view from the eye in red dotted lines.

Further, there are a number of shapes with unique characteristics such as those in Figure 4a-c. The first could appear on its side or on its back end with the protruding elements acting as stabilizers, so it might be tilted towards the camera. The second could easily be stud-side down, on each vertical side, on the slant, but not likely stable on its bottom or backside. The third might have 5 stable sides. In other words, the number of possible cells of an aspect partition differs for each brick type. It is quite possible to enumerate all of these given we have the library of part labels; however, that enumeration would be tedious, and an approximation will suffice. We assume, as mentioned earlier, that each unique brick has an average of 3 stable poses, so for 1130 unique brick types there would be a total of 3390 stable brick appearances due to changing pose. Call these expected stable poses $p_1$, $p_2$, and $p_3$ (perhaps right side up, upside down, and on the left side, to pick one possible set of examples). For our cube example, above, a missing face would impact 9 out of 26 aspects, or about a third of all aspects. In other words, the 'left out' pose in $A^p$ must be one of the brick's stable poses, $p_1$, $p_2$, or $p_3$, and not an arbitrary pose.

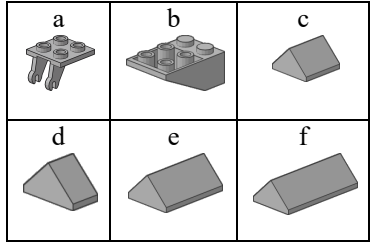

**Figure 4.** Examples of bricks for Section 4.4.

What is the impact if one stable pose, say $p_1$, of all brick types being left out of a training set? The aspect graph itself would be incorrect. Not only would one face (at least) be missing (or presumed flat) but its interactions with its neighboring faces would be incorrect. Recognition of that particular brick, if based on the learned visible aspects, would be impaired unless the particular viewpoint the camera sees is one from which only properly learned aspects are visible. If a third of the aspects are affected, then assuming all viewpoints are equiprobable, one third of all views of this brick would lead to erroneous recognition.

It is worth pointing out that the blocks in Fig. 4c-f are examples of objects for which degenerate views are possible. The cross-section of the each is the same, only the length differs, so even though a block maybe in a stable pose, the imaging geometry would yield an ambiguous situation (Dickinson et al. 1999).

Is it possible that the other aspects for this brick or the aspects of other bricks, could together combine to permit correct recognition? We could consider this on a case-by-case basis, but this would be quite tedious. Nevertheless, it might be, with generous assumptions about the capabilities of the learned network, that the learned elements corresponding to the bricks in Figure 4d-f could combine to provide what is needed to recognize the slope of intermediate size, shown in the previous set. But there is no corresponding brick combination for the first one in the previous set, nor for many other bricks; they are all quite unique. Thus, recognition of those bricks is assumed to fail if the proper constellation of aspects is not seen.

It seems safe to think that the error would be larger than that of the ideal network which has already been pegged at $G_A < 0.000018$. If no brick is seen with $p_2$, for example, it means that one third, or 1130, of the possible brick appearances would have to be recognized as a result of generalization. We can make generous assumptions about the generalization capabilities of the network, specifically that it can successfully handle similar bricks such as the roof-like ones just depicted even if no training sample for a particular pose is present. However, a rough count in the brick library yields 132 unique pieces that have no similar ones from which a generalization can easily obtained, or in other words, 12% of the possible 1130 unique bricks.

A different possibility would be some kind of data augmentation in training set preparation. No data augmentation could remedy missing poses because no such method inserts patches out of other images (the stud surface appears in other poses of course). Even a data augmentation that could consider geometry would not be able to fill in the missing samples unless prior knowledge of what all the invisible aspects could look like, or assumptions of surface continuity, is somehow applied in the data augmentation process (see also Logan et al. 2018, Ian et al. 2014). In any case, such an action has its own problems (Rosenfeld et al. 2018) and would not be a sensible strategy.

Our hypothetical network $A^{-p}$ would have the following error, assuming the least error implied by the above arguments. Suppose that the generalization process is effective for the bricks for which there might be additive ways of deriving a brick (even if not straightforward), but for the 132 unique bricks they are all classified incorrectly, a 12% error as described. Thus, an estimate of overall test error is $G_{A^{-p}} < 0.120018$. This estimate is likely too small.

## 4.5 Training Set Missing One Brick Shape

Suppose one shape is left out and that the resulting system $A^{-s}$ satisfies the empirical error stipulated above. Shape is a more abstract notion here than the other dimensions considered. It is included because one might imagine a customer requesting an order of all sloped bricks, of all rectangular bricks, of all plates, etc. For example, the bricks in Figure 5a-d are all considered plates and their regularity is the thickness of the base. On the other hand, the bricks in Figure 5e-h are classified as slopes because they all contain a sloped surface.

There are 106 slope brick sub-types. The question here is if all plates (or slopes, or bricks, or wedges, etc.) were left out of the training set, could the resulting network generalize so that they could be recognized sufficiently well?

Let us consider some of the details of our assumed network $A^{-s}$. These images contain no texture information useful to the task, thus we assume that none is learned. The imaging geometry, especially the lighting is such that shadows or other cues for shape from shading are not useful. We can also leave out color for the current purpose. It is well-known that many learned networks represent receptive field tunings in early layers very similar to oriented Gabor filters in their early layers and this seems a good first level of processing for the brick images. One can easily imagine what a line drawing of each of the LEGO® bricks might be; the bricks are textureless. It is an acceptable assumption then that processing then continues on such a representation.

How can shape be inferred from a line drawing? There is a wealth of literature on how computers might interpret line drawings. LEGO® brick shapes are mostly polyhedra and one of their shape characteristics (but not a complete characterization to be sure) is the set of labels of their lines and vertices. A labelling of an image is an assignment to each edge of the image of one of the symbols +, -, ⇢ and ⇠ (concavity or convexity), and similarly for each vertex one of the many types of vertices (Clowes 1971, Waltz 1971; several sources enumerate the set, but the exact cardinality is not important). Such a labelling is a reasonable assumption as a representation of shape sufficient to enable discrimination of one shape from another, although it is not difficult to show counterexamples. For the purposes of our argument, this does not really cause any difficulties. Such an 'in principle' solution is instructive even if not precisely what a trained neural network realizes.

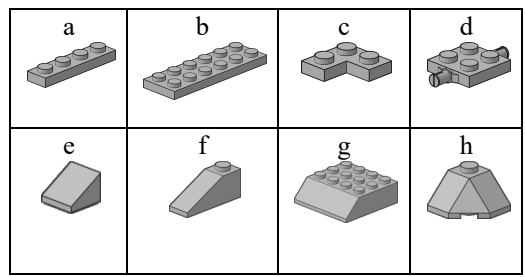

**Figure 5.** First set of example bricks for Section 4.5.

Kirousis & Papadimitriou (1988) examined the processing of images of straight lines on the plane that arise from polyhedral scenes. They asked, given an image, is there a scene of which the image is the projection? Such images are called realizable. One classical approach to the realizability problem is through a combinatorial necessary condition. A labelling is legal if at each node of the image there is only one of the legal patterns. A legal labelling is consistent with a realization of the image, if the way the edges are seen from the projection plane is the way indicated by the labelling. They provided a proof that it is NP-complete, given an image, to tell whether it has a legal labelling. This is true for opaque polyhedra, and is even true in the simpler case of trihedral scenes (no four planes share a point) without shadows and cracks. Although there are methods for labelling such a line drawing, the problem is that it is exponential to determine if the labelling corresponds to a real object. In other words, since any algorithm for extracting straight lines from images necessarily may have error, any error may signal an illegal labelling where there is none or a legal labelling where there is not one, so a step requiring the verification is needed. Once it is known that there is a legal labelling, there exist algorithms for matching labelled image to known objects that have known labellings. These results, importantly, are independent of algorithm or implementation; they apply to the problem as a whole.

Kirousis & Papadimitriou (1988) also present an effective algorithm for the important special case of orthohedral scenes (all planes are normal to one of the three axes). It is tempting to think that this latter case applies to LEGO® bricks; most are indeed approximately orthohedral (the studs pose the exception), but then again there are many pieces within the library that are strictly polyhedral or involve curved segments. For example, the class Plates 1xN includes the bricks of Figure 6a, the class Plates 2xN includes the bricks of Figure 6b. Slopes include the bricks of Figure 6c. The class Brick 1xN includes bricks such as that of Figure 4d. Even in decomposition these will contain elements that are non-polyhedral.

The task of realizing a general, non-orthohedral scene, given its labelled image can be solved by employing linear programming, i.e., a polynomial time algorithm exists. So, the problem remains to find the legal labelling. It is known that the labeling of trihedral scenes is NP-complete as is the complexity of labeling origami scenes, that is, scenes constructed by assembling planar panels of negligible thickness (Parodi 1996, Sugihara 1982, Malik 1987).

It is difficult to accept that a deep learning procedure can effectively learn the solution to an NP-Complete problem; it might, however, learn approximate solutions that are within some error bound $\varepsilon^{-s}$ for subsets of the full problem. We will not explore this route but suggest that this makes the generalization issue even more difficult to address.

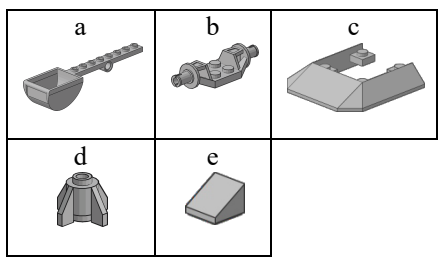

**Figure 6.** Second set of example bricks for Section 4.5.

Note that we still have not come to the generalization issue. Suppose no slopes are included in the training set but a customer wishes a box of all slopes in the LEGO® catalog. This means that no instance of the configuration of lines and vertices seen in bricks such as that in Figure 6e have participated in any learning. It is highly unlikely that a network can construct a particular configuration out of learned elements that would suffice; after all, as shown, the problem in general is combinatorial and even generous assumptions about learned networks cannot defeat this fact.

In order to quantify the expected error of the hypothetical network $A^{-s}$, we can be optimistic and say that the error would not be greater than the error incurred by mis-classifying the smallest group of bricks in the catalog. In other words, if all slopes are erroneously classified because slopes were not in the training set, it would mean that the remaining 520 - 106 = 414 types are correctly classified. The smallest such group in the catalog is that of plates with 22 instances. Then, if those 22 classes are not part of the training set, 22 x 6 (poses) x 6 (colors) x 72 (orientations) = 9504 images are mis-classified out of the possible 1,468,800 images, or 0.647%. Thus, an estimate of overall test error is $G_{A^{-s}} < 0.006488$. It should be clear that this is a very optimistic estimate.

## 4.6 Interim Summary

Having looked at 5 different cases of selection bias in training sets, we summarize the above analysis in Table 2. The hypothetical networks $G_{A^{-c}}$, $G_{A^{-z}}$, $G_{A^{-p}}$ and $G_{A^{-s}}$ all have orders of magnitude too great an error given the production standards. As should be clear, if the training set is biased with respect to any characteristic except orientation in the imaging plane or non-primary colors, the resulting network is unlikely to generalize so that the required performance standard can be met. The particular values for generalization error may invite argument to be certain. However, there are two indisputable facts that emerge. First, the error for these four networks is far beyond the acceptable manufacturing values; it's not small nor negligible. Second, there are at least a few training biases that cannot be generalized away. LEGO® bricks are a simple domain, one that can be completely characterized. Imagine how such a thought experiment might be impacted by a more complex domain.

| Training Set Bias | Learned Network | Generalization Error |
|---|---|---|
| Unbiased | $A$ | $G_A < 0.000018$ |
| Color Biased | $A^{-c}$, c ∈ (red, blue, yellow) | $G_{A^{-c}} < 0.166685$ |
|  | $A^{-c}$, c ∈ (green, black, white) | $G_{A^{-c}} < 0.000018$ |
| Size Biased | $A^{-z}$ | $G_{A^{-z}} < 0.018602$ |
| Orientation Biased | $A^{-o}$ | $G_{A^{-o}} < 0.000018$ |
| 3D Pose Biased | $A^{-p}$ | $G_{A^{-p}} < 0.120018$ |
| Shape Biased | $A^{-s}$ | $G_{A^{-s}} < 0.006488$ |

**Table 2.** Summary of training bias analyses.

# 5.0 General Discussion

The previous sections attempt to show that generalization in learned networks may not be what it seems. We will first address potential issues regarding our assumptions, approach, and analysis methods. Then, we relate our thought experiment to real domains, specifically, autonomous driving, in order to show its relevance.

## 5.1 Methodology and Assumptions

We have sought to answer the question *If a network is trained with a dataset that misses particular values of some defining domain attribute, can it generalize to the full domain from which that training dataset was extracted while maintaining its performance accuracy?* This is different than what many others have considered regarding the generalization behavior of learned networks. Rather than conduct a deep mathematical analysis or extensive empirical analysis neither of which has provided an answer, we tried a different approach, a Thought Experiment. The strategy we used is Proof by Contradiction via Backcasting, both well-known and widely used methods. Backcasting involves establishing the description of a very definite and very specific future situation. This specific situation involved the assumption that our question was in fact answered in the positive, that learned networks effectively generalize even when training biases are present and that they effectively bridge such biases. We then defined a real, yet simple, domain on which we would test this future situation, that of LEGO® bricks. This seems to be a novel twist to the overall problem because it ties real engineering performance bounds on the network's performance, something not often seen in the previous literature.

Regarding the details, we made a number of assumptions involving the LEGO® domain and since all were made to represent the lower magnitudes of any counts, the overall set of such assumptions can be considered generous towards a negative result to our thought experiment. We assumed that the learned systems $A$ and $A^{-i}$ work perfectly and that training error equals zero. We also assumed that regardless of the training regimen or network configuration, each network includes all the classes that would be required to compose any user query (specifically, classes for each brick sub-type, each pose type, each orientation, each color).

It is certainly true that in vision, no current training set fully captures all dimensions of the representation of visual information. So, this has direct relevance to computer vision systems, and likely beyond to other domains. Our thought experiment unequivocally points to the conclusion that any generalization behavior is completely dependent on the acceptable system error and on the nature of the domain data not represented in the training set.

Finally, it would be easy to argue that for the LEGO® domain, the space of possible images is small enough so that a well-designed neural network would perform very well, or even that there is now enough computer power cheaply available to enable a brute-force search solution. Both these points are true and in practice perhaps this is exactly how LEGO® might develop a practical system for this purpose. It is just as valid to think that there could be many engineering workarounds or manipulations for problems with how object pose or orientation might appear on a conveyor belt. Examples abound on modern manufacturing lines (tracking guides, slots, use of gravity, friction or puffs of air for positioning, etc.). But this was not the point of our exercise. We had no desire to present a realizable design for a production system. The point was to select a real and complex task, whose scale is manageable for complete representation in order to test the limits of generalization in the face of training biases and performance requirements.

## 5.2 Cases Outside the Brick Binning Task

The argument of Section 4.0 assumes that only one value of some dimension is missing. What if the training bias encompasses some combination of attributes? For example, for some bricks a particular color may be missing while for others some orientation is missing. Or perhaps there are no red bricks in the training set and some 3D poses are also missing. Worse still, suppose there are no yellow bricks and no slopes in the training set. It is straightforward to devise many more combinations. For some the impact might be small while for others large. It seems apparent that there are many combinations where no apparent satisfactory behaviour would be observed in terms of LEGO®'s production standards.

The LEGO® brick application was discussed with a few assumptions to simplify the hypothetical system. In a different real world setting when arbitrary kinds of objects might be involved in normal scenes, these assumptions will not be valid. First of all, there would not be a single well-isolated object in a scene; there would be many, likely with interactions among them. Changes in viewpoint of the observer or in general, the overall geometry of the image formation process (Faugeras 1993), shadows or highlights and how they change with illumination direction and surface properties (Horn 1986), accidental alignments of object features (Dickinson et al. 1999), and more will need to be considered. A detailed analysis of each of these complexities is beyond our scope; however, just for illustrative purposes, we consider the viewpoint change case for a set of assembled bricks, an object more complex than a single brick but still not quite as complex as an arbitrary real-world object.

Suppose we are presented with a construction of 'standard' LEGO® bricks (all edges are orthogonal to the others), say on a LEGO® plate of 10 x 24 studs and the plate is presented with some random 3D pose to the observer. Let us ignore for the moment the impact of the studs on the brick surfaces. A simple enough task for an observer might be simply count the number of bricks used in the construction. For our hypothetical scenario above, suppose a customer wanted 1000 random constructions of this form each with exactly $K$ bricks. Assuming a random brick construction engine, these items might pass under the same camera on a conveyor belt and a system to count the bricks would then decide which random construction to give to this particular customer. As mentioned earlier, Kirousis & Papadimitriou (1988) provide an effective algorithm for labelling all the visible lines and vertices of orthohedral scenes so the recognition of each visible brick for counting purposes (only) is assumed solved. However, it is not difficult to see that if a brick is occluded by another, some or all of its lines and vertices would not be labelled and thus it could not be counted unless the viewpoint changes. The incidence of occlusions depends on the stable poses of the brick construction. One, perhaps the most, stable pose is for the supporting plate to be flat on the conveyor belt. Occlusions would still occur if the construction included bricks that are partially supported and this overhang perhaps above other bricks. Two other stable positions can be formed where one of the the long ends (there are 2 long ends) of the plate is on the belt but the overall construction is tipped over with some other contact point and the edge forming a stable pose. Here, occlusions would be frequent. As in the case of the previously described network $A^{-p}$, where one 3D pose is left out of the training set, no amount of training could fill in the structure that cannot be observed. Active observation has been shown to be critical for many real-world visual tasks (Andreopoulos & Tsotsos 2013, Bajcsy et al. 2018). It is not difficult to imagine a wide variety of tasks outside the ones mentioned where similar problems would arise. It is important to note that for many, solutions could be easily engineered to avoid the problems; but that was not the point of our analysis. Our point was to ask questions about generalization and in that respect, we conclude that great care needs to be applied when making assumptions about how a trained network can or cannot generalize.

## 5.3 Real-World Implications

It is important to address why our analysis is representative of how generalization might appear in other, real-world, visual domains and tasks. The first point to make is that it is clear that the creation of a complete image dataset is an impossibility regardless of domain and thus any real training set will include gaps if not also outright biases. Generalization is necessary. Suppose that by some crazy quirk, the training set for an autonomous driving scenario has no images where a traffic light is yellow, only green or red and that there are no yellow cars, no yellow yield signs, nothing yellow at all. As we showed above, the system will not generalize in order to ever recognize yellow. The point here is that without a careful analysis of exactly what is and is not in the training set, one can never be certain of generalization behaviour in a safety-critical situation.

Rasouli & Tsotsos (2019) survey in a very broad fashion the issues relevant for how autonomous vehicles might interact with pedestrians. This is, obviously, only one component of the full autonomous driving task but will suffice to make our point here. They show that there are at least 22 major factors, and among these, at least 12 require visual perception and decision-making. Among these 12 are included pedestrian gender, age, road structure, traffic volume, road signal systems, and so on. The variety of appearances of road signal systems alone is daunting. For our thought experiment approach, one might think of a dataset that, due to no fault of its creators, might not represent young children as pedestrians very well, or might not represent all skin tones well (a problem well-noted for the face recognition task). Thus, it is easy to see how one might proceed with the thought experiment in the autonomous driving setting and reach much the same conclusions.

The number of accidents involving autonomously driven automobiles is increasing and here we give a couple of examples to illustrate how important generalization should be. Consider the accident in which a person walking a bike was killed[9]. The system creators may have indeed had a category for pedestrian and a category for bicycle (either stationary or being ridden) but may not have a category for person-walking-a-bike (across the street). An NTSB 2018 study concluded that the self-driving system first registered radar and LIDAR observations of the pedestrian about 6 seconds before impact. The vehicle speed was 43 mph. As the vehicle and pedestrian paths converged, the system classified the pedestrian as unknown, then as a vehicle, and then as a bicycle with varying path projections. At 1.3 seconds before impact, the self-driving system initiated emergency braking in order to avoid a collision. First, hypothetically there might be the possibility that person-walking-a-bike is never a targeted category for the detection system such that it has difficulty appropriately registering the situation as an unknown object, as a vehicle, and then as a bicycle. Secondly, it is unclear that the detection NN could compose a person-walking-a-bike out of detections of a person and simultaneously a more-or-less co-located and co-travelling bike without being explicitly engineered to do so.

Consider a second example. A video of an autonomous vehicle crashing directly into an overturned box truck in Taiwan went viral, partly because it showcases what many assume to be a failure of the vehicle's autopilot feature[10]. The clip makes it apparent the truck should be an easy obstacle to detect and avoid if one is paying attention to the road. It is hard to miss a giant white blockade in the middle of the highway. The autopilot nevertheless presumably failed to detect the truck, the driver was forced to put on the brakes himself, but it was already too late to avoid the collision. There was also a sign 100 metres before the overturned truck alerting any driver to the crash, which means that both driver and autopilot system ignored that, too. Was this an error of generalization? Perhaps. The system could not have been trained on all possible images it might see. It would depend on its ability to generalize across such training gaps. An overturned white truck in the middle of a path should have been an easy one for a human visual system if it were part of the current field of view (Section 4.4 showed how a missing viewpoint impacts network performance; this could be such a case). It might also be the case that this was not a simple matter of pattern recognition; rather, a more complex kind of generalization might be at play, one that requires reasoning about consequences of acting within the observed environment.

There is an important reminder from such tragedy: neural networks and their underlying theory provide no guarantee that a given network will be able to cover all the relevant dimensions or categories or compositions unless they are included and tested accordingly. Human perception and cognition do not depend solely on classification of images. Carroll (1993) gives scores of human cognitive abilities as examples. Humans see but also reason and act (see Bajcsy et al. 2018). It may be that a new unification of classical methods in computer vision and AI with current machine learning methods will yield the best of both.

## 5.4 Limits on Generalization

It should be clear that no neural network can make a provably undecidable problem decidable, nor a provably intractable problem tractable (such as the NP-Complete problems described in Section 4.5). For such problems, the only recourse is approximation. The main question then is how far such an approximation would be from the production requirements of some particular application, such as our hypothetical brick binning task. For example, in the missing pose case of Section 4.4, it would not be unexpected that a full 3D brick might be classified even if the network was not trained on one possible pose; an interpolation would be likely that fills in any gaps. But how good is that interpolation in practice. The same is true for the missing color case of Section 4.1; it is certain that some color mixture might be output by the network, but how close is it to the required one? Modern applications employing learned networks for some domains approach high-90% accuracy levels, quite an accomplishment to be sure. However, our brick binning task needs several of orders of magnitude better performance. Many applications involving human life and safety require even lower error bounds. In other applications, such as face recognition for photo-sharing, or casual language translation, errors have virtually no associated costs (from a life and safety perspective), so our analysis may seem not so relevant. Further, the proofs in Kratsios & Bilokopytov (2020) and others make it appear as if the approximation error can be arbitrarily small (but non-zero). However it seems a great challenge to approach this theoretical limit given how the reported empirical errors of the latest neural networks are many orders of magnitude higher than those for our brick binning task. These are empirical questions that cannot be answered here but are certainly important for future study.

---

[9] https://www.nytimes.com/2018/03/19/technology/uber-driverless-fatality.html

[10] https://driving.ca/tesla/auto-news/news/tesla-model-3-presumed-to-be-on-autopilot-plows-into-overturned-truck

## 5.5 The Problems with Data Augmentation

As has been mentioned a few times so far, the set of techniques collectively known as data augmentation, a data-space solution to the problem of limited data, are commonly used to supplement the data assembled for training sets in learned systems (for reviews see van Dyk & Meng, 2001, Shorten & Khoshgoftaar, 2019, among others). Such techniques enhance the size and quality of training datasets leading to better systems. This is done under the assumption that more information can be extracted from the original dataset through augmentations. These augmentations artificially inflate the training dataset size by either data warping or oversampling. Many other strategies for increasing generalization performance focus on the model's architecture itself leading to progressively more complex architectures (Shorten & Khoshgoftaar, 2019). Other specific functions for this problem include functional dropout regularization, batch normalization, transfer learning, and pretraining (e.g., Kukačka et al., 2017, Hernández-García & König 2018). This paper focuses on visual data; the image augmentation algorithms include geometric transformations, color space augmentations, kernel filters, mixing images, random erasing, feature space augmentation, adversarial training, generative adversarial networks, neural style transfer, and meta-learning (Shorten & Khoshgoftaar, 2019). The augmented data is intended to represent a more complete sampling of possible data points, thus decreasing the distance between the training and validation set, as well as test sets.

All of these have proved themselves empirically to some degree but they all feature 3 issues. First, these are not principled approaches with respect to the actual data sampling deficiencies. Second, the data augmentation methods have secondary effects which seem ignored but which are potentially detrimental. Finally, they are not targeted at a domain's performance specification, they simply seek to improve generalization in some neutral manner. Consider the first of these issues.

Any data set that purports to adequately represent some domain in order for a learning system to extract the relevant statistical regularities important for that domain, must actually include those features in a statistically discoverable manner. All of the methods increase the size of the training data via manipulations of the existing data in the belief that the enhanced data set better samples the domain. It is puzzling that these are not based on an analysis of the data set in order to discover exactly what might be missing, that is, what features are not statistically well represented.

Consider next the issue of secondary effects. The augmented image set may help generalization, but the added samples have the potential secondary effect of changing the balance of representation for other features. Take a simple example. In order to increase rotation invariance as a performance criterion, an existing image set might be augmented by rotated versions of those images (see Logan et al. 2018, where this is also discussed). The artificially rotated images however change the distribution of lighting source directions across the full set of images. This may not be relevant for many applications, and of course humans have little difficulty with this in general (but are not immune to it; e.g., 'hollow face illusion', Hill & Bruce 1993). But for those applications where the learned system is part of an agent that needs to make inferences about lighting direction (to interpret shadows or depth cues from surface shading, etc.), this has the potential of improving performance in one variable while damaging that of another as a secondary effect. The same would be true for augmentation via symmetry manipulations; again the imaging geometry changes. Another manipulation involves simple spatial shifts in order to enhance position invariance or image padding to ameliorate boundary effects. This changes the context within which target objects are found and as seen in Rosenfeld et al. (2018), context is very important and can have quite detrimental impact on accuracy of classification. Learning methods learn not only the required target objects but also the context within which they are found. Much more on such effects appears in Shorten & Khoshgoftaar (2019).

These secondary effects will have a ripple effect across the full set of features a domain requires. Each manipulation although targeted to some specific feature, necessarily changes sampling of other feature dimensions. Without targeting direct knowledge of the training set sampling deficiencies, the third issue mentioned above, data augmentation may improve global performance in an indirect manner and not necessarily performance along a particular feature dimension of importance. We should also point out that these image manipulation methods rarely go out of the 2D image plane; objects in the real world are three-dimensional and none of these methods can possibly augment in the third dimension let alone how 3D imaging geometry (light sources, camera viewpoint and settings, object position relative to illumination and camera, ambient or reflected illumination, surface properties) affect appearance (although see Ian et al. 2014 where the assumptions of sufficient view data and object surface continuity seems to permit new views, but not with accompanying illumination changes).

## 5.6 Bias in Datasets

In Section 1.2, the experimental biases described by Sackett (1979) were very briefly discussed, ending with the suggestion that these, with suitable domain translations, might be as valid for computer vision systems (as well as AI in general) as for medical data. Certainly the potential for bias is very real. Consider the number of possible training sets in the visual domain. Recall the estimate of number of discernable images presented earlier, $1.5 \times 10^{25}$. For the sake of argument, suppose that the number of samples used for training is $10^7$ (the scale ImageNet represents as mentioned earlier). How many subsets of $10^7$ images are possible? This is given by $\frac{1.5 \times 10^{25}!}{10^7! \, (1.5 \times 10^{25} - 10^7)!}$, not so easy to evaluate, but certainly impossibly large to be certain which of these possible training sets are free of bias. Although it might be easy to search in a brute-force manner for a training set that yields incrementally better performance numbers than others, this is not the point. Bias in that training set remains unaddressed. Debate over Pavlidis' estimate will not really lead to a better result because even if that estimate is reduced to $10^{10}$, the number of possible subsets of a billion elements remains insanely large[11].

Even so, many shrug off such combinatorial arguments as not relevant. After all, humans are able to understand any image they see on the internet and many feel that machines should also have this ability. Such a casual conclusion arises from a lack of knowledge; this ignores the fact that humans do this at different speeds and with different performance levels (see Carroll (1993) for a comprehensive presentation of human visuospatial abilities; see Itti, Rees & Tsotsos (2005) for an encyclopedic treatment of the breadth of characteristics seen in human and primate attentional behavior, among others). Reaching such a casual conclusion is entirely unjustified and if it forms a pillar of the empirical method for the field, there is an important need for re-evaluation. Certainly, the phrase 'sees like humans' which is very frequently attached to descriptions of modern computer vision research needs to be re-considered. This is especially true if the computer system outperforms humans. To see like a human, from an external observer's point of view, should mean that the accuracy is similar, that the time course to a decision is similar, that the number and kinds of errors made are similar, and that any additional external behavior (eye movements for example) are similar. In general, these are not reported, and it is unlikely these are true for modern systems. It is a great goal to build a machine that can 'see like humans', but one must be clear about what this actually entails. By eye alone, humans may not match engineering performance specifications such as those described here.

It seems intractable and infeasible to attempt an empirical examination of all potential sources and impacts of bias when developing a training and test set of data. However, as we have shown, a procedure like our thought experiment might be a way to begin addressing the issue. Tied to a domain-specific performance target, it seems straightforward to extend the reasoning described here in order to determine the impact of many types of bias. This might be time-consuming and tedious, but it is tractable. Rasouli & Tsotsos (2019) have shown that it is possible to develop a substantial (not necessarily complete but very useful nonetheless) set of factors relevant to how drivers deal with pedestrians. A similar analysis could be done for other domains.

---

[11] For amusement, try any of the online permutation/combination calculators (e.g., https://www.calculator.net/permutation-and-combination-calculator.html) and ask how many combinations without replacement there are of 1,000 elements out of 10,000; the count has 1,410 digits!

It may be useful to require that the creation of any training or test set be accompanied by a *Bias Breakdown*, that is, an analysis of the spectrum of possible biases that is explicit regarding which biases are relevant and which are not, which can be dealt with by good sample choices and which can not, and what any impacts on final system performance might be due to the remaining biases.

This *Bias Breakdown* can be organized easily. One suggestion for a template could be the following. For each source of bias - the 56 Sackett describes might be too much, but it's a good starting point - one might complete this table:

| *Bias class* | *Bias dimensions evaluated* | *Interpretation within current domain* | *Bias not applicable by design* | *Evaluated impact on performance requirements* | *Bias dimensions not evaluated* | *Estimated impact of non-evaluated biases* |
|---|---|---|---|---|---|---|
| | | | | | | |

In our thought experiment, as already described, we focussed on Selection Bias, and within Sackett's sub-classes, *missing data bias* fits best. Completing this single dimension of the table yields:

| *Bias class* | missing data |
|---|---|
| *Bias dimensions evaluated* | size, shape, orientation, color, 3D pose |
| *Interpretation within current domain* | there are no training samples representing a particular value of one or more of these dimensions |
| *Bias not applicable by design* | the following are assumed constant across all training/test samples: imaging geometry, lighting, occlusion |
| *Evaluated impact on performance requirements* | See Table 2 above |
| *Bias dimensions not evaluated* | conjunctions of size, shape, orientation, color, 3D pose; low sample size of some dimension |
| *Estimated impact of non-evaluated biases* | less predictable:<br>- no samples of multiple dimensions likely will increase error<br>- low sample sizes of a single dimension may lead to unpredictably less error<br>- low sample sizes of multiple dimensions likely to increase error above that for lowest error on single dimension with zero samples |

This should be considered a starting point and experience will determine how such a *Bias Breakdown* may evolve. Note that it is important to know for each application what the acceptable performance requirements might be and to to tie the goal of generalization to it. For safety-critical applications, the need for a clear, as complete as feasible, documentation of biases in training data and their impacts (or biases in any other aspect of system development, similar to the manner Sackett describes bias across the full research enterprise) cannot be over-emphasized.

## 6.0 Conclusions

This thought experiment was motivated by the desire to understand the nature of generalization in learned computer vision systems. At some basic level, this understanding should not depend on the exact characteristics of the system architecture or learning method. We sought to understand generalization at a more fundamental level. Specifically, we considered the impact of Selection Bias on generalization, because in vision, a training set is a very tiny subset of the full population of all possible images. We employed a well-understood methodology in order to challenge the limits of generalization, Reductio ad absurdum (proof via contradiction) via Backcasting. Our analysis indeed found contradictions. These were mostly due to physics, to geometry or to computational intractability for which no amount of cleverness in an algorithm can compensate. Sampling bias may be hidden and silent. It is hidden because very few research efforts verify that a domain is appropriately sampled to create the training set, and silent because the errors it causes are seen only if appropriate test cases are used (even though many have tried to raise the point, e.g., Akhtar & Mian 2018). Any learning system is agnostic as to what is missing when trained, whether it be values of attributes or whole attribute dimensions. One needs to be methodical and principled in dataset construction, statistically valid regarding sampling with respect to all the variations of the data, data augmentation and generalization expectations. Our analysis also found that generalization errors would exceed, by orders of magnitude, the engineering requirements of our toy (yet not so 'toy') domain. Our thought experiment points to three conclusions: first, that generalization behavior is dependent on how sufficiently the particular dimensions of the domain are represented during training; second, that the utility of any generalization is completely dependent on the acceptable system error; and third, that specific visual features of objects, such as pose orientations out of the imaging plane or colours, may not be recoverable if not explicitly represented sufficiently in a training set. It may be that more principled design of not only training sets but of the whole empirical protocol for learned systems is necessary. Current learned systems and their underlying theory provide no guarantee that a given system will be able to cover all the relevant dimensions, categories or compositions unless they are included and tested accordingly. It may be that a new unification of classical methods in computer vision and AI with current machine learning methods, with each paying more attention to the variability caused by object pose, viewpoint, and more generally the full imaging and lighting geometry, will yield the best of both. Whereas empirical confirmation may be the best method for confirming generalization effectiveness, it is a time and resource expensive activity. Our *Thought Experiment Probe* approach, coupled with the resulting *Bias Breakdown*, is far less costly and can be very informative as a first step towards understanding the impact of biases. Further, it is can be important supplementary documentation component for training data intended for systems with critical performance requirements.

## Acknowledgements
The authors wish to thank Ershad Banijamali, Iuliia Kotseruba, Amir Rasouli, Amir Rosenfeld, Brian Cantwell Smith, Sven Dickinson and Konstantine Tsotsos for their helpful comments and suggestions on earlier drafts of this manuscript.